\documentclass[10pt,a4paper,twocolumn]{article}
\usepackage[margin=18mm,columnsep=7mm]{geometry}
\usepackage[T1]{fontenc}
\usepackage{lmodern,microtype,amsmath,amssymb,booktabs,array,graphicx,xcolor,hyperref}
\usepackage{tikz,pgfplots}
\usepackage{float} % keep the reference-parameter table adjacent to its discussion
\pgfplotsset{compat=1.18}
\usetikzlibrary{arrows.meta,positioning,decorations.pathmorphing}
\hypersetup{colorlinks=true,linkcolor=blue!55!black,urlcolor=blue!55!black,citecolor=blue!55!black}
\title{\bfseries Model Predictive Control of a Multirotor UAV with a Cable-Suspended Payload: A Comparison of Shifted and Transformer-Based Warm Starts}
\author{Athanasios Zeris\thanks{Independent Researcher, Athens, Greece. \mbox{Email: \href{mailto:athzeris@gmail.com}{athzeris@gmail.com}}. ORCID: \href{https://orcid.org/0009-0002-6907-2400}{0009-0002-6907-2400}.}}
\date{}
\begin{document}\raggedbottom\maketitle
\begin{abstract}
Multirotor drones carrying cable-suspended payloads require coordinated regulation of vehicle motion and suspended-load oscillations while respecting rotor constraints and cable dynamics. In this study, the drone is modeled as a multirotor unmanned aerial vehicle (UAV). This study develops a physics-based model predictive control (MPC) framework for a multirotor with an elastic cable and a reduced double-pendulum representation of its suspended load, and compares two approaches to initializing its sequential convex programming (SCP) optimizer: conventional receding-horizon shifting and Transformer-based sequence prediction. The controller combines numerical propagation of a coupled mechanical model with move-blocked control decisions, finite-difference sensitivities, constrained quadratic subproblems, and feasibility projection of the initial control sequence. Expert SCP solutions are used to train the Transformer, and the learned and shifted initializations are evaluated using the same prediction model, constraints, and solver settings. Mechanical consistency checks, data-generation and feasibility studies, tests of independently solved MPC problems, and sequential closed-loop experiments establish a common basis for the comparison. Although Transformer initialization improves computational performance relative to zero initialization in the tests of independently solved MPC problems, the shifted solution is more efficient in the sequential closed-loop benchmark and achieves comparable tracking behavior. The results show that learned sequence accuracy and downstream optimizer efficiency are distinct properties, and that previous-solution shifting is an essential baseline when assessing learning-assisted MPC. The measured optimization times remain above the nominal control sampling period, so the study does not demonstrate real-time deployment.
\end{abstract}

\section*{Nomenclature}
\noindent Symbols are defined below; numerical parameter values are reported separately in Table~\ref{tab:params}.
\par\medskip\noindent\textbf{Coordinates and physical parameters}\par\smallskip
\noindent\begingroup\footnotesize\renewcommand{\arraystretch}{1.03}
\begin{tabular}{@{}p{0.33\columnwidth}p{0.60\columnwidth}@{}}
\toprule \textbf{Symbol} & \textbf{Definition} \\ \midrule
$q\in\mathbb R^{11}$ & Generalized coordinates \\
$\dot q$ & Generalized velocities \\
$x=[q^\top,\dot q^\top]^\top$ & Mechanical state ($22$ components) \\
$r_Q,r_B,r_P,r_L$ & UAV reference, attachment, intermediate-mass and terminal-load positions \\
$R$ & Body-to-inertial rotation matrix \\
$J_Q,J_P,J_L$ & Translational Jacobians \\
$J_\omega$ & Angular-velocity Jacobian \\
$M(q)$ & Generalized mass matrix \\
$G(q)$ & Gravity-force vector \\
$c(q,\dot q)$ & Velocity-dependent generalized force \\
$m_Q,m_P,m_L$ & UAV, intermediate and terminal masses (kg) \\
$\rho$ & Cable length (m) \\
$\ell_c,\ell_P$ & Cable rest length and secondary-link length (m) \\
$k_c,c_c$ & Cable axial stiffness (N/m) and damping (N s/m) \\
$\alpha,\beta,\alpha_2,\beta_2$ & Cable and payload swing angles (rad) \\
$g$ & Gravitational acceleration (m/s$^2$) \\
$T_c$ & Cable tension (N) \\
\bottomrule
\end{tabular}\endgroup\par
\par\medskip\noindent\textbf{MPC and control allocation}\par\smallskip
\noindent\begingroup\footnotesize\renewcommand{\arraystretch}{1.03}
\begin{tabular}{@{}p{0.33\columnwidth}p{0.60\columnwidth}@{}}
\toprule \textbf{Symbol} & \textbf{Definition} \\ \midrule
$u=[T,\tau_\phi,\tau_\theta,\tau_\psi]^\top$ & Virtual manipulated inputs: thrust and body moments \\
$f=[F_1,F_2,F_3,F_4]^\top$ & Individual rotor thrusts (N) \\
$\Gamma$ & Rotor allocation matrix ($u=\Gamma f$) \\
$u_h,x_h$ & Hover input and reference state \\
$y$ & Controlled-state residual components \\
$d$ & External disturbance (not injected in matched tests) \\
$T_s$ & Control sampling period (s) \\
$N_p,N_b$ & Prediction samples and control blocks \\
$j$ & Prediction-step index ($j=1,\ldots,N_p$) \\
$p_j,\ p_j^{\rm ref}$ & Predicted and desired UAV positions at step $j$ (m) \\
$\dot p_j$ & Predicted UAV translational velocity at step $j$ (m/s) \\
$e_j$ & Predicted 17-component tracking-error vector at step $j$ \\
$S$ & Virtual-input scaling matrix \\
$z\in\mathbb R^{12}$ & Normalized blocked control decisions \\
$\mathcal Z$ & Admissible decision set \\
$A,b$ & Linear inequality coefficients for rotor constraints \\
$J$ & Finite-horizon objective \\
$r,w$ & Weighted residual and state-error weights \\
\bottomrule
\end{tabular}\endgroup\par
\par\medskip\noindent\textbf{SCP and learned initialization}\par\smallskip
\noindent\begingroup\scriptsize\renewcommand{\arraystretch}{1.0}
\begin{tabular}{@{}p{0.33\columnwidth}p{0.60\columnwidth}@{}}
\toprule \textbf{Symbol} & \textbf{Definition} \\ \midrule
$D_i$ & Finite-difference residual Jacobian at iteration $i$ \\
$H_i$ & Regularized Gauss--Newton Hessian approximation \\
$g_i$ & Local objective gradient \\
$\delta$ & SCP decision increment \\
$\Delta$ & Trust-region radius \\
$\Pi_{\mathcal Z}$ & Projection onto admissible decision set \\
$\pi_\theta$ & Transformer decision predictor \\
$T_{\rm total}$ & Inference, projection and SCP runtime (s) \\
\bottomrule
\end{tabular}\endgroup\par

\section{Introduction}
Multirotor unmanned aerial vehicles (UAVs) carrying suspended loads are coupled mechanical systems: changes in rotor thrust and vehicle attitude affect the payload trajectory, while cable deflection, extension, and payload swing feed back into the forces acting on the vehicle. A controller intended for constrained tracking must therefore represent the relevant coupled dynamics without making repeated online prediction prohibitively expensive. This balance between mechanical fidelity and computational effort motivates the present comparison of model predictive control (MPC) initialization strategies.

The suspended-load literature offers different levels of mechanical detail. Goodarzi, Lee, and Lee \cite{goodarzi2014} derive a coordinate-free quadrotor--payload formulation with a flexible cable represented by serially connected links and develop a geometric stabilization controller with stability analysis. Tolba and Shirinzadeh \cite{tolba2025} instead consider a fifteen-degree-of-freedom multirotor and rigid-body payload model that includes elastic cable behavior, slackening and tightening, shifted centers of gravity, arbitrary attachment locations, and double-pendulum motion; they investigate an output-weighted linear-quadratic regulator. These studies establish the importance of cable and payload dynamics, but neither is an evaluation of Transformer initialization for a common predictive optimizer. Our control-oriented model has eleven generalized coordinates, including vehicle translation and attitude, an extensible suspension segment, and two swing-direction pairs. It does not reproduce all the effects represented in the more detailed models, and the reported results must be interpreted within that modeling scope.

Predictive control provides a framework for handling tracking objectives and explicit constraints \cite{rawlings,dingyang2024,borrelli2017}. For quadrotors, Ru and Subbarao \cite{ru2017} formulate constrained trajectory-tracking MPC using a state-dependent-coefficient representation and sequential quadratic programming. Rohr et al. \cite{rohr2021} develop and flight-test predictive velocity control for a tiltwing UAV using an identified aerodynamic model; their use of preceding solutions as optimization references should be distinguished from using a shifted solution as the initial iterate. These works illustrate the importance of selecting a control-oriented prediction model and a practical numerical optimization procedure, but do not directly establish the performance of the coupled suspended-payload controller considered here.

Computational cost remains a central issue in receding-horizon control. Wang and Boyd \cite{wangboyd} demonstrate that exploiting quadratic-program structure can substantially accelerate online MPC. Move blocking reduces the number of independent control decisions, but Cagienard et al. \cite{cagienard} also show why blocking choices cannot be assumed to preserve stability and constraint-satisfaction guarantees. Diehl, Bock, and Schl\"oder \cite{diehl} develop a real-time iteration scheme for approximate online optimization, providing a complementary example of trading iterative solution accuracy against the time available between control updates. In the present sequential convex programming (SCP) implementation, the numerical workload also includes repeated integration of the coupled mechanical model to construct finite-difference sensitivities and evaluate trial steps. Faster quadratic-program solutions alone therefore need not imply faster control updates.

Learning-based models address different parts of this computational problem. Gu, Primatesta, and Rizzo \cite{gu2024} investigate physics-informed neural modeling of quadrotor dynamics, highlighting model accuracy and physical consistency; their network serves a different purpose from the initialization predictor studied here. Davoudi et al. \cite{davoudi2020} likewise illustrate the influence of aerodynamic and wind-model fidelity on quadrotor flight simulation. Our Transformer does not replace the mechanical prediction model or estimate the wind field. It predicts an initial control-decision sequence for subsequent physics-based SCP optimization.

Transformer-assisted trajectory optimization provides the immediate methodological motivation. Celestini et al. \cite{celestini} use sequence modeling to generate initial guesses or target plans for constrained trajectory optimization and report improvements in convergence and runtime in their evaluated settings. Such benefits cannot be presumed for a different plant, solver, and prediction horizon. In particular, a shifted previous solution is an inexpensive and informative baseline in receding-horizon control, and learned inference and feasibility projection add their own computational costs.

We therefore compare three initializations of the \emph{same} move-blocked SCP controller: zero deviations from hover, a shifted and reblocked previous solution, and a Transformer-generated candidate. The shifted and Transformer-generated initial control sequences are adjusted, when necessary, to satisfy the imposed input constraints before optimization. Zero initialization already satisfies these constraints. The reported experiments assess both isolated optimization behavior and matched closed-loop episodes, including measured initialization-and-optimization time, solver iterations, and physical tracking outcomes. The principal contribution is a controlled numerical comparison of zero, shifted, and Transformer-based initialization strategies within a common SCP-MPC framework for a coupled UAV--cable--payload system. The study evaluates their effects on computational effort, optimization iterations, and closed-loop response under identical modeling and control constraints. The results establish the relative performance of these initialization strategies for the investigated numerical scenarios, without implying formal closed-loop stability or real-time flight readiness.

\begin{figure*}[t]
\centering
\resizebox{\textwidth}{!}{%
\begin{tikzpicture}[>=Latex, font=\sffamily\footnotesize,
box/.style={draw=blue!50!black,rounded corners=2pt,fill=blue!4,align=center,minimum height=10mm,inner sep=4pt},
method/.style={draw=violet!65!black,rounded corners=2pt,fill=violet!6,align=center,minimum height=10mm,inner sep=4pt},
scp/.style={draw=teal!65!black,rounded corners=2pt,fill=teal!7,align=center,minimum height=10mm,inner sep=4pt},
arrow/.style={->,thick,draw=black!70}]
% Physical system: original AI-assisted technical sketch,
% with separately drawn two-stage suspended system matching the manuscript.
\node[font=\bfseries] at (1.6,3.2) {Coupled UAV--payload model};
\node[inner sep=0,anchor=center] at (1.6,2.50)
  {\includegraphics[width=3.15cm]{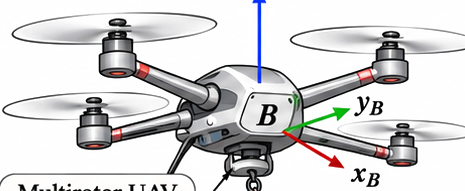}};
% The UAV sketch is supplemented by the two-stage suspension below
% to match the 11-DOF model, without changing other panels.
\draw[very thick,orange!80!black,decorate,decoration={zigzag,segment length=4pt,amplitude=.8pt}]
  (1.62,1.82) -- (1.47,1.20);
\shade[ball color=gray!50] (1.47,1.08) circle (.13);
\draw[thick,gray!70!black] (1.47,.95) -- (1.75,.50);
\shade[ball color=gray!60] (1.75,.40) circle (.20);
\node[font=\scriptsize,anchor=west] at (1.67,1.13) {$m_P$};
\node[font=\scriptsize,anchor=west] at (2.0,.40) {$m_L$};

\node[box,minimum width=2.2cm] (state) at (4.6,2.2) {Current state $x_k$\\and tracking reference};
\draw[arrow] (2.7,1.9) -- (state.west);
\node[method,minimum width=2.7cm] (shift) at (8,3.0) {Shift previous solution\\one sample; reblock};
\node[method,minimum width=2.7cm] (tr) at (8,1.55) {Transformer prediction\\first 12 of 60 outputs};
\draw[arrow] (state.east) -- ++(.35,0) |- (shift.west);
\draw[arrow] (state.east) -- ++(.35,0) |- (tr.west);
\node[box,minimum width=2.5cm] (proj) at (11.35,2.25) {Feasibility projection\\$\Pi_{\mathcal Z}$};
\draw[arrow] (shift.east) -- ++(.32,0) |- (proj.west);
\draw[arrow] (tr.east) -- ++(.32,0) |- (proj.west);
\node[scp,minimum width=2.6cm] (opt) at (14.75,2.25) {Shared SCP optimizer\\physics rollout + QP\\trust region / line search};
\draw[arrow] (proj.east) -- (opt.west);
\node[box,minimum width=2.5cm] (act) at (14.75,.45) {Apply first control\\advance by $T_s$};
\draw[arrow] (opt.south) -- (act.north);
\draw[arrow] (act.west) -- (4.6,.45) -- (state.south);
\node[draw=gray!65,fill=gray!5,rounded corners,align=center,inner sep=5pt] at (8.2,-.65) {Common matched configuration: $T_s=0.20$ s; $N_p=6$ samples; $N_b=3$ blocks of 2 samples; $z\in\mathbb R^{12}$\\Zero initialization is a reference baseline; shift and Transformer are alternative warm starts, not simultaneous inputs.};
\end{tikzpicture}}
\caption{Physical system and matched receding-horizon SCP-MPC architecture. The two warm-start branches are compared separately using the same feasibility projection and SCP solver. The quadrotor is illustrated using a conceptual sketch; the two-stage suspension is schematic. The prediction model and numerical dimensions are those of the reported numerical benchmark.}\label{fig:architecture}
\end{figure*}

\begin{figure*}[t]
\centering
\resizebox{.80\textwidth}{!}{%
\begin{tikzpicture}[>=Latex, node distance=7mm, font=\sffamily\footnotesize,
  proc/.style={draw=blue!55!black,fill=blue!4,rounded corners,align=center,text width=8.1cm,minimum height=9mm,inner sep=4pt},
  init/.style={draw=violet!65!black,fill=violet!5,rounded corners,align=center,text width=3.5cm,minimum height=12mm,inner sep=4pt},
  arr/.style={->,thick}]
\node[proc] (input) {Read $x_k$, reference, and previous solution};
\node[init,below left=10mm and 0mm of input] (sh) {Shift previous solution and reblock};
\node[init,below right=10mm and 0mm of input] (tr) {Transformer prediction; retain 12 decisions};
\draw[arr] (input.south) -- ++(0,-3mm) -| (sh.north);
\draw[arr] (input.south) -- ++(0,-3mm) -| (tr.north);
\node[proc,below=23mm of input] (proj) {Project the selected initialization onto the admissible control set $\mathcal Z$};
\draw[arr] (sh.south) |- (proj.west);
\draw[arr] (tr.south) |- (proj.east);
\node[proc,below=of proj] (roll) {Physical rollout, residuals, finite-difference Jacobian};
\node[proc,below=of roll] (qp) {Solve constrained trust-region QP};
\node[proc,below=of qp] (line) {Feasibility check, line search, iterate update};
\node[proc,below=of line] (stop) {Converged or four iterations?};
\node[proc,below=of stop] (apply) {Apply first control; advance plant by $T_s=0.2$ s};
\foreach \a/\b in {proj/roll,roll/qp,qp/line,line/stop,stop/apply}{\draw[arr] (\a.south)--(\b.north);}
\draw[arr] (stop.east) -- ++(8mm,0) |- (roll.east) node[pos=.2,right]{continue};
\end{tikzpicture}}
\caption{Algorithmic flow of the matched SCP--MPC comparison. The shift and Transformer branches are alternatives, not simultaneous inputs. Zero initialization is a separate reference baseline. The projection enforces input-decision constraints, not general state-path safety.}\label{fig:algorithm}
\end{figure*}
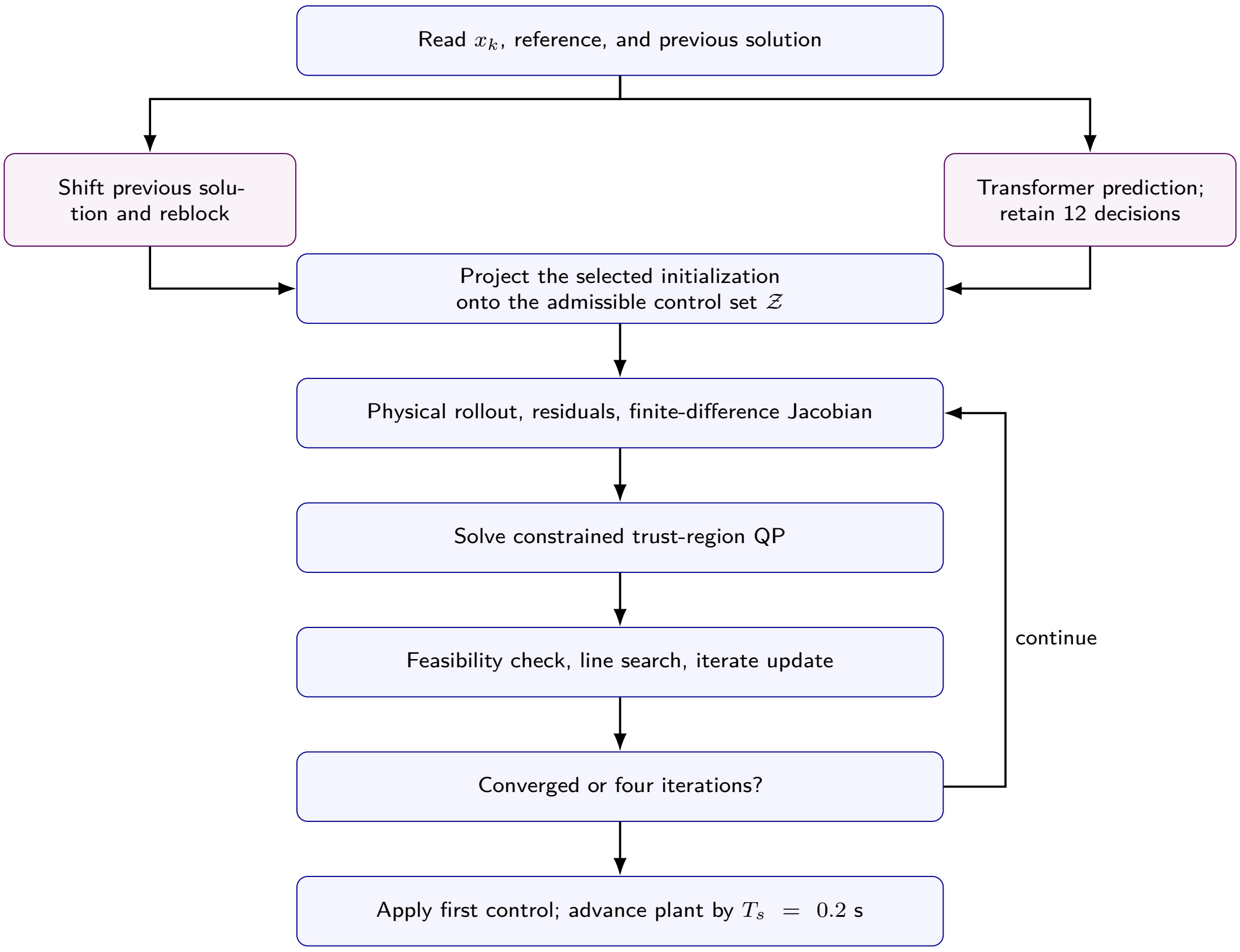

\section{Mechanical model and prediction}
The model has eleven generalized coordinates and twenty-two states, $x=[q^\top,\dot q^\top]^\top$. Its equations can be represented as
\begin{equation}
M(q)\ddot q+f_{\rm internal}(q,\dot q)=Q(q,u),\qquad \dot x=F(x,u).
\end{equation}
The implemented dynamics include the multirotor, elastic cable, and suspended payload. To check stationary hover, the validation script computed the norm of the 11 generalized accelerations, $\|\ddot q_h\|_2=\|\dot x_h(12{:}22)\|_2$. The result, approximately $8.86\times10^{-10}$, indicates numerical equilibrium consistency. Because the coordinates include translational and angular variables, this is not a force residual in newtons. At hover, the cable tension was 4.905 N and each rotor produced about 4.574 N of thrust. These numerical checks are not flight-test validation. The coordinate definitions, mass-matrix construction, elastic-cable force, and force allocation are documented below directly from the implemented source.

The controller samples every $T_s=0.2$ s and predicts $N_p=6$ steps (1.2 s). The SCP predictor uses guarded fourth-order Runge--Kutta integration with 0.02-s substeps; closed-loop plant propagation uses an adaptive explicit Runge--Kutta method of orders 4 and 5, with a maximum step of 0.04 s. The two integrations are intentionally distinct.

\subsection{Generalized coordinates and geometric construction}
The source defines the coordinates in the fixed order
\begin{equation}
q=[x,y,z,\phi,\theta,\psi,\alpha,\beta,\rho,\alpha_2,\beta_2]^\top.
\end{equation}
The inertial coordinate system is $z$-up; body orientation is represented by a ZYX Euler rotation $R=R_z(\psi)R_y(\theta)R_x(\phi)$. The first three coordinates give the UAV reference position $r_Q$. The cable attachment point, intermediate suspended mass, and terminal load positions are
\begin{align}
r_B &=r_Q+Rr_{B/A},\\
r_P &=r_B+\rho\,d(\alpha,\beta),\\
r_L &=r_P+\ell_P\,d(\alpha_2,\beta_2),\\
d(a,b)&=[\sin a\cos b,\ \sin b,\ -\cos a\cos b]^\top.
\end{align}
The cable length coordinate $\rho$ is a state, not a fixed kinematic constant. The model also contains a second suspended-body direction, so the two pairs of swing angles describe distinct components of the coupled system. These definitions follow the archived kinematics implementation.

\subsection{Mass matrix, gravity, and generalized inputs}
The fast dynamics source forms the mass matrix from the translational Jacobians of the UAV, intermediate mass, and terminal load, together with the UAV rotational inertia:
\begin{equation}
M(q)=m_QJ_Q^\top J_Q+m_PJ_P^\top J_P+m_LJ_L^\top J_L+J_\omega^\top J_Q^{\rm body}J_\omega.
\end{equation}
Here $J_Q,J_P,J_L$ are the position Jacobians of $r_Q,r_P,r_L$ with respect to $q$, and $J_\omega$ maps Euler rates to body angular velocity. The code symmetrizes $M$ numerically. The gravity term is assembled as
\begin{equation}
G(q)=g\sum_{a\in\{Q,P,L\}}m_a J_a(3,:)^\top.
\end{equation}
The generalized thrust and moment contributions are $Q_{1:3}=R[0,0,u_1]^\top$ and $Q_{4:6}=E(\phi,\theta)^\top[u_2,u_3,u_4]^\top$. Thus the optimization variables are collective thrust and body moments, mapped to four rotor thrusts through the allocation matrix $\Gamma$. The phrase ``four rotor-control channels'' in the abstract refers to these four virtual inputs, not four independently optimized rotor forces.

\subsection{Elastic cable, gyroscopic terms, and numerical implementation}
The axial cable force is unilateral and uses the implemented piecewise rule
\begin{equation}
T_c=\begin{cases}
\max\{0,k_c(\rho-\ell_c)+c_c\dot\rho\},&\rho\geq\ell_c,\\
0,&\rho<\ell_c.
\end{cases}
\end{equation}
The generalized radial force includes $-T_c$. The velocity-dependent generalized force is evaluated using the Christoffel contraction associated with the mass matrix:
\begin{equation}
\begin{aligned}c(q,\dot q)&=\dot M(q)\dot q-\tfrac12\nabla_q(\dot q^\top M(q)\dot q),\\M(q)\ddot q&=Q-c-G.\end{aligned}
\end{equation}
This term represents the Coriolis and centrifugal effects arising from the configuration dependence of the mass matrix. The mass-matrix derivatives are evaluated numerically using centered finite differences. The fast RHS computes eleven mass-matrix directional derivatives using $10^{-5}$ coordinate perturbations. It checks the ZYX Euler singularity condition $|\cos\theta|<0.1$ and solves the mass-matrix system at each evaluation. The nominal experiments set the optional generalized damping vector to zero in the closed-loop controller setup.

\subsection{Reference parameters and verification}
The archived parameter file attributes reference parameter values to the study in \cite{tolba2025}; for example, the rotor arm length is set to 0.2107~m.
\begin{table}[H]\centering\small\renewcommand{\arraystretch}{1.4}
\caption{Selected physical parameters used in the archived model.}\label{tab:params}
\begin{tabular}{@{}lll@{}}\toprule
Parameter & Value & Unit\\\midrule
UAV mass $m_Q$ & 1.365 & kg\\
Intermediate suspended mass $m_P$ & 0.300 & kg\\
Terminal load mass $m_L$ & 0.200 & kg\\
Cable rest length $\ell_c$ & 0.25 & m\\
Cable stiffness $k_c$ & 500 & N/m\\
Cable damping $c_c$ & 3.5 & N s/m\\
Secondary suspended-body length $\ell_P$ & 0.20 & m\\
Rotor arm length & 0.2107 & m\\
Maximum rotor thrust & 9.842 & N\\
Gravitational acceleration & 9.81 & m/s$^2$\\\bottomrule
\end{tabular}
\end{table}
This is an adopted model input, not a value obtained by optimization or calibration in the present work. The published reference system has fifteen degrees of freedom and a rigid-body suspended payload, whereas our implemented model has eleven generalized coordinates and a reduced two-stage load representation. They should not be regarded as identical mechanical models. The hover consistency check, free-oscillation integration, and mass-matrix conditioning are numerical checks, not flight-test validation.\par
In a separate two-second free-oscillation pilot, the simulation started from the hover equilibrium with the initial cable and payload swing angles set to $0.05$~rad and $-0.03$~rad, respectively. The rotor inputs were held at their hover values, while the UAV was allowed to move. The computed maximum absolute angular excursions were approximately $2.865^\circ$ for the cable and $4.576^\circ$ for the payload; the simulated cable length varied from 0.259720 to 0.259914~m. These numbers are simulation outputs, not prescribed angle limits or experimental measurements, and they are separate from the closed-loop MPC benchmark.
\section{Model predictive control formulation and sequential convex programming}
\subsection{Control objectives, feedback, and references}
The \emph{manipulated variables} are four virtual commands $u=[T,\tau_\phi,\tau_\theta,\tau_\psi]^\top$: collective thrust and body moments. The physical actuators are rotor thrusts $f=[F_1,F_2,F_3,F_4]^\top$, related by $u=\Gamma f$. The exact source-code allocation (rotors numbered as in the parameter file) is
\begin{equation}
\Gamma=\begin{bmatrix}
1&1&1&1\\
0&L&0&-L\\
-L&0&L&0\\
\kappa&-\kappa&\kappa&-\kappa
\end{bmatrix},\quad\begin{gathered}L=0.2107\,{\rm m},\\\kappa=0.01503.\end{gathered}
\end{equation}
The feedback signal is the complete simulated state $x_k\in\mathbb R^{22}$ (ideal full-state feedback); the experiments do not implement a sensor/observer pipeline. The reference is the model's hover equilibrium $x_h$: position $(0,0,0.8)$ m, zero Euler and swing angles, zero velocities, and cable extension $\rho_h=\ell_c+(m_P+m_L)g/k_c=0.25981$ m. The nominal input is $u_h=[(m_Q+m_P+m_L)g,0,0,0]^\top=[18.29565,0,0,0]^\top$ in SI units. The \emph{controlled quantities} are the weighted position, attitude, swing angles, selected velocities, and control effort specified in Section~3.3. No external disturbance is injected in the final matched comparison: $d(t)=0$ in the modeled tests. In particular, no claim of wind rejection or measurement-noise robustness follows from this benchmark.
\begin{table}[t]
\centering\footnotesize
\caption{MPC variables and assumptions in the matched closed-loop benchmark.}\label{tab:mpc-definition}
\renewcommand{\arraystretch}{1.15}
\begin{tabular}{@{}p{0.36\columnwidth}p{0.57\columnwidth}@{}}
\toprule
\textbf{Category} & \textbf{Implemented definition}\\\midrule
Manipulated variables & Virtual collective thrust and three body moments, $u=[T,\tau_\phi,\tau_\theta,\tau_\psi]^\top$; mapped to four rotor thrusts $f$ \\
Setpoints & Hover position $(0,0,0.8)$ m; zero attitude, swing angles, and velocities; hover cable extension $\rho_h=0.25981$ m \\
Feedback / measured variables & Ideal simulated full state $x_k\in\mathbb R^{22}$; no explicit sensor or observer model \\
Controlled quantities & Position, attitude, cable/payload swing, and selected velocities in the weighted tracking residual; control effort and block changes penalized \\
Disturbances & No injected wind or other external disturbance in the matched benchmark ($d(t)=0$); no measurement noise \\
Prediction and decisions & $T_s=0.2$ s; $N_p=6$ samples (1.2 s); three two-sample blocks; $z\in\mathbb R^{12}$ \\
Input limits & Normalized decision box bounds and rotor-thrust constraints via $u=\Gamma f$ \\
\bottomrule
\end{tabular}
\end{table}

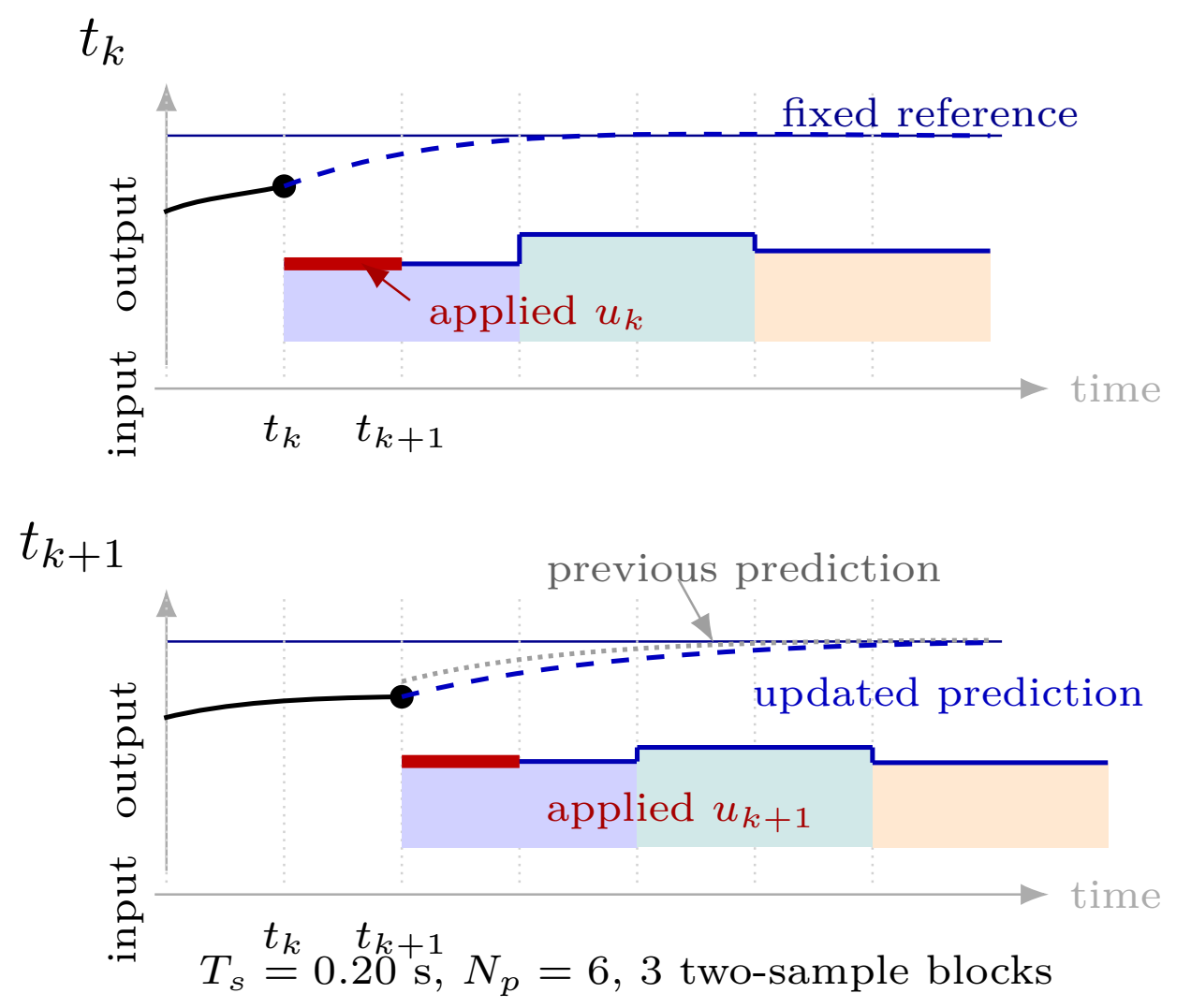
\begin{figure}[t]
\centering
\resizebox{\columnwidth}{!}{%
\begin{tikzpicture}[x=0.70cm,y=0.70cm,font=\scriptsize,>=Latex]
  % Consecutive optimization panels, same time scale
  \foreach \base/\lab in {4.65/{t_k},0.35/{t_{k+1}}}{
    \draw[gray!65,->] (-.1,\base)--(7.5,\base) node[right]{time};
    \draw[gray!65,->] (0,\base+.20)--(0,\base+2.60);
    \node[anchor=east,font=\bfseries] at (-.14,\base+2.95) {$\lab$};
    \node[anchor=east,rotate=90] at (-.34,\base+2.0) {output};
    \node[anchor=east,rotate=90] at (-.34,\base+.52) {input};
    \draw[blue!55!black,thin] (0,\base+2.15)--(7.1,\base+2.15);
    \foreach \j in {0,...,6}{\draw[gray!35,densely dotted] (\j,\base+.1)--(\j,\base+2.55);}
  }
  % Top: state and predicted trajectory (schematic)
  \draw[black,thick] (-.02,6.15) .. controls (.28,6.27) and (.55,6.28) .. (1,6.37);
  \fill[black] (1,6.37) circle (2pt);
  \draw[blue!75!black,dashed,thick] (1,6.37) .. controls (2,6.75) and (3.1,6.79) .. (4,6.81) .. controls (5.2,6.82) and (6.4,6.80) .. (7,6.80);
  % Reference for top at y=6.80; label
  \node[anchor=west,blue!55!black] at (5.05,7.00) {fixed reference};
  % Top blocks: t_k=1, next sample t_k+1=2
  \foreach \a/\b/\shade/\height in {1/3/blue!18/.66,3/5/teal!18/.91,5/7/orange!18/.77}{
    \fill[\shade] (\a,5.05) rectangle (\b,5.05+\height);
    \draw[blue!70!black,thick] (\a,5.05+\height)--(\b,5.05+\height);
  }
  \draw[blue!70!black,thick] (3,5.71)--(3,5.96) (5,5.96)--(5,5.82);
  \draw[red!75!black,line width=2.2pt] (1,5.71)--(2,5.71);
  \node[red!65!black,anchor=west] at (2.05,5.29) {applied $u_k$};
  \draw[red!65!black,->] (2.07,5.40)--(1.65,5.72);
  \node[anchor=north] at (1,4.60) {$t_k$};
  \node[anchor=north] at (2,4.60) {$t_{k+1}$};
  % Bottom: new measured state, old vs updated prediction
  \draw[black,thick] (-.02,1.85) .. controls (.5,1.98) and (1.15,2.02) .. (2,2.03);
  \fill[black] (2,2.03) circle (2pt);
  \draw[gray!75,densely dotted,thick] (2,2.16) .. controls (3,2.42) and (4.4,2.51) .. (7,2.51);
  \draw[blue!75!black,dashed,thick] (2,2.03) .. controls (3,2.29) and (4.5,2.46) .. (7,2.49);
  \node[anchor=west,gray!75!black] at (3.05,3.10) {previous prediction};
  \draw[gray!75,->] (4.35,3.03)--(4.65,2.49);
  \node[anchor=west,blue!75!black] at (4.8,2.04) {updated prediction};
  \foreach \a/\b/\shade/\height in {2/4/blue!18/.73,4/6/teal!18/.85,6/8/orange!18/.72}{
    \fill[\shade] (\a,.75) rectangle (\b,.75+\height);
    \draw[blue!70!black,thick] (\a,.75+\height)--(\b,.75+\height);
  }
  \draw[blue!70!black,thick] (4,1.48)--(4,1.60) (6,1.60)--(6,1.47);
  \draw[red!75!black,line width=2.2pt] (2,1.48)--(3,1.48);
  \node[anchor=west,red!65!black] at (3.05,1.05) {applied $u_{k+1}$};
  \node[anchor=north] at (1,.29) {$t_k$};
  \node[anchor=north] at (2,.29) {$t_{k+1}$};
  \node[anchor=west] at (.1,-.34) {$T_s=0.20$ s, $N_p=6$, 3 two-sample blocks};
\end{tikzpicture}}
\caption{Original schematic of receding-horizon MPC at consecutive sampling instants. The predicted output is updated using the newly available state, and only the first control sample is applied at each update. Colored intervals denote three two-sample move blocks; the displayed output and input levels are illustrative rather than recorded data. The shift--reblock operation is illustrated separately in Fig.~\ref{fig:shift}.}\label{fig:mpc-timeline}
\end{figure}

The receding-horizon operation is illustrated in Fig.~\ref{fig:mpc-timeline}. At time $t_k$, the controller predicts six future samples, optimizes the three blocked inputs, and applies only the first sample. At $t_{k+1}=t_k+T_s$, the state is updated and the optimization is solved again. The shifted and Transformer-based methods change the initial candidate for this repeated optimization, not its physical prediction model or input constraints.

\subsection{Prediction and move blocking}
Four normalized control deviations are optimized over $N_b=3$ blocks, each lasting two prediction samples. With hover input $u_h$ and scaling $S=\operatorname{diag}(2,0.15,0.15,0.05)$,
\begin{equation}
u_{j|k}=u_h+Sz_{b(j),k},\quad b(j)=\lceil j/2\rceil,\quad z_k\in\mathbb R^{12}.
\end{equation}
The four control components are total rotor thrust and three body moments, which have different units and numerical ranges. The scaling matrix $S$ expresses their adjustments about hover through normalized decision variables: a unit change in the corresponding component of $z$ represents a change of 2~N in total thrust, 0.15~N$\,$m in either of the first two moments, or 0.05~N$\,$m in the third moment. These are the ranges implied by the normalized box bounds, not necessarily the final allowable input ranges; the rotor-thrust constraints can impose tighter limits. The scaling values are fixed controller settings, not parameters identified or optimized in this study, and are identical for all three initialization methods.
\subsection{Objective function and input constraints}
The normalized three-block decisions $z$ are shared by all initialization methods, not only shifting. The feasible decision set is $\mathcal Z=\{z:Az\le b,\ -\mathbf 1\le z\le\mathbf 1\}$, where $A,b$ encode the rotor-thrust bounds through the allocation matrix. A finite-horizon residual $r(z;x_k)$ collects weighted position, attitude, cable/payload swing, and velocity deviations, as well as control and control-move penalties. The implemented objective is
\begin{equation}
J(z;x_k)=\tfrac12\|r(z;x_k)\|_2^2.
\end{equation}
At prediction step $j$ ($j=1,\ldots,N_p$), the predicted UAV position $p_j\in\mathbb R^3$ is compared with its desired position $p_j^{\rm ref}$. The desired position is fixed at $(0,0,0.8)$~m in the reported tests. The tracking-error vector also includes UAV attitude, cable and payload swing angles, and selected velocities; their reference values are zero. With the prediction step indexed by $j$ rather than the hover subscript $h$, its 17 components are
\begingroup\renewcommand{\arraystretch}{1.2}
\begin{equation}
\label{eq:tracking-error}
 e_j=\begin{bmatrix}
 p_j-p_j^{\rm ref}\\
 \phi_j\\\theta_j\\\psi_j\\
 \alpha_j\\\beta_j\\\alpha_{2,j}\\\beta_{2,j}\\
 \dot p_j\\
 \dot\alpha_j\\\dot\beta_j\\\dot\alpha_{2,j}\\\dot\beta_{2,j}
 \end{bmatrix}\in\mathbb R^{17}.
\end{equation}
\endgroup
The current MPC update index $k$ is omitted here for readability; the weighted stage residual is $w\odot e_j$. The residual's physical-state weight vector is $(30,30,50,2,2,1,8,8,8,8,1,1,2,1,1,1,1)^\top$; the control-effort and successive-block-difference residual coefficients are $\sqrt{0.04}$ and $\sqrt{0.10}$, respectively. Terminal guidance is disabled in the matched closed-loop benchmark.

\subsection{Sequential convex programming solution}
At iteration $i$, finite differences produce the residual Jacobian $D_i$. The regularized Gauss--Newton approximation to the Hessian is $H_i=D_i^\top D_i+10^{-4}I$, while the corresponding gradient is $g_i=D_i^\top r_i$. The resulting quadratic subproblem is
\begin{equation}
\min_d\;\tfrac12d^\top H_id+g_i^\top d
\quad\text{s.t.}\quad A d\le b-Az_i,
\quad \underline d_i\le d\le\overline d_i.
\end{equation}
The step bounds include a trust radius of 0.15 and the original box bounds. A backtracking line search accepts an improving feasible trial. The maximum SCP iteration count is four; finite-difference step size is $10^{-3}$. A valid final decision need not satisfy the code's convergence stopping criterion.

\subsection{Implementation audit of the matched controller}
The MPC formulation was checked against the archived closed-loop simulation code and its saved results. The controller uses a sampling time of $T_s=0.2$~s, predicts six future samples, and groups them into three blocks of two samples. It therefore optimizes 12 decision variables, $z\in\mathbb R^{12}$. The SCP algorithm performs at most four iterations, uses a trust-region radius of 0.15, a finite-difference perturbation of $10^{-3}$, and a regularization term of $10^{-4}I$.

Both the initial-candidate projection and the local quadratic subproblem are solved using an interior-point convex quadratic-programming solver. The Jacobian of the residual is estimated using a forward finite difference when the forward evaluation is valid. If it is not valid, the code uses a backward finite difference instead. It is not generally a central-difference Jacobian.

A five-level backtracking line search accepts a trial step only if it is feasible and reduces the objective. The MPC predictor uses guarded fourth-order Runge--Kutta integration with 0.02-s substeps. The simulated plant uses an adaptive fourth/fifth-order Runge--Kutta integrator.

These checks confirm that the manuscript accurately describes the archived implementation. They do not constitute an independent re-execution or validation of the simulations.

\subsection{Why the model-based SCP stage remains computationally demanding}
Each candidate control sequence requires the 22-state system to be integrated over six prediction intervals. To construct the finite-difference Jacobian, the code perturbs each of the 12 control-decision variables separately. Therefore, each local quadratic approximation requires several additional physical-model rollouts, beyond the trial evaluations used by the line search.

As a result, the main computational cost is not necessarily the quadratic-programming subproblem itself. It is the repeated physical prediction required to evaluate candidate controls and estimate sensitivities. The optimization therefore consists of a sequence of model rollouts and QP solves. A learned initial guess can reduce computation only if it reduces the number of these rollouts or QP solves, or their associated costs.

\subsection{Input feasibility versus closed-loop safety}
Before SCP begins, the shifted or Transformer-generated initial candidate is projected onto the admissible control-decision set, which includes the rotor-thrust constraints and normalized box bounds. During SCP, the same input limits are imposed in every local quadratic subproblem, and the returned decision is checked for feasibility.

These checks establish that the recorded solutions satisfy the implemented input limits. They do not guarantee collision avoidance, payload safety, cable-tension limits, robustness to disturbances, closed-loop stability, or recursive feasibility. A feasible control sequence may still produce excessive payload swing, excessive cable tension, or unsafe motion.

In these numerical experiments, a control decision is counted as valid when it satisfies the imposed input constraints and improves the objective. This is different from satisfying the SCP convergence test: an acceptable control decision may be obtained even when the optimizer stops before convergence. Input feasibility alone does not establish safe UAV--payload motion, closed-loop stability, or robustness to disturbances.

\section{Initialization strategies}
\subsection{Zero initialization}
The zero-initialization baseline sets the normalized control-decision vector to $z_k^{(0)}=0$ at every sampling instant. Because the decision variables represent adjustments about the hover input $u_h$, this choice starts the optimization from the nominal hover command over all three control blocks. It does not mean that the physical rotor thrusts are zero. The initial vector already satisfies the implemented input bounds, so no projection is required. SCP then computes the control sequence using the same prediction model, objective, and constraints as for the shifted and Transformer initializations.
\subsection{Shifted warm start with fixed move blocking}
\begin{figure}[ht]\centering
\begin{tikzpicture}[x=0.85cm,y=0.65cm, every node/.style={font=\small}]
\foreach \i/\lab/\col in {0/A/blue!25,1/A/blue!25,2/B/teal!25,3/B/teal!25,4/C/violet!25,5/C/violet!25}{\draw[fill=\col] (\i,2) rectangle +(0.93,0.72);\node at (\i+0.46,2.36){\lab};}
\foreach \i/\lab/\col in {0/A/blue!25,1/B/teal!25,2/B/teal!25,3/C/violet!25,4/C/violet!25,5/C/violet!25}{\draw[fill=\col] (\i,0.7) rectangle +(0.93,0.72);\node at (\i+0.46,1.06){\lab};}
\node[anchor=east] at (-0.2,2.36){Previous};\node[anchor=east] at (-0.2,1.06){Shifted};
\draw[-{Stealth}] (3,1.93)--(3,1.48);
\node[align=center] at (3,-0.15){Reblock: $(A+B)/2\quad (B+C)/2\quad C$};
\end{tikzpicture}
\caption{One-sample horizon shift followed by reblocking into three fixed two-sample blocks. The averaged sequence is projected onto the admissible control-decision set.}\label{fig:shift}
\end{figure}
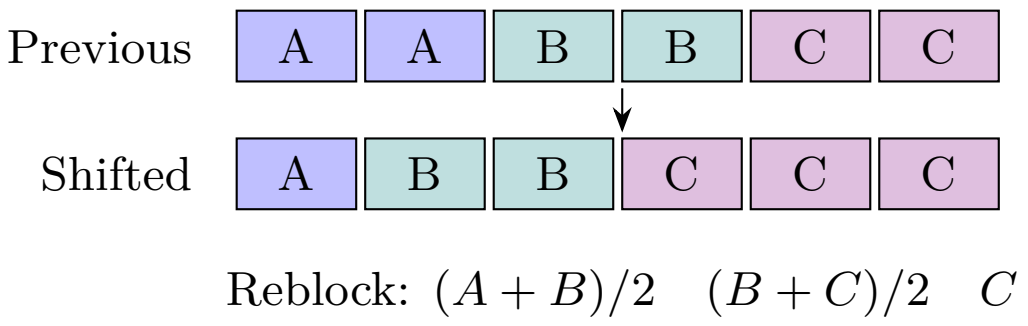

Let the previous optimized blocks be $Z_k^*=[a,b,c]$, with $a,b,c\in\mathbb R^4$. Expanding the blocks gives $[a,a,b,b,c,c]$. Advancing the horizon by one sample and holding the terminal action gives $[a,b,b,c,c,c]$. Since the next optimizer still uses three fixed two-sample blocks, pairwise averaging yields
\begin{equation}\label{eq:shift}
\widetilde Z_{k+1}=\left[\frac{a+b}{2},\frac{b+c}{2},c\right],\qquad
z_{k+1}^{(0)}=\Pi_{\mathcal Z}(\operatorname{vec}\widetilde Z_{k+1}).
\end{equation}
This is a shifted warm start \emph{adapted to fixed move blocking}, not a controller with moving block boundaries. At the first update, when no prior solution exists, it starts from zero. Figure~\ref{fig:shift} depicts the operation.

\subsection{Transformer-based initial control sequence}
At each sampling instant, the current state $x_k$ is supplied to the trained network. Its output is a candidate sequence of future virtual control adjustments. This sequence is used only to initialize the SCP iterations; the subsequent control optimization uses the same plant model, objective, and constraints as the zero and shifted cases.

The network was trained to predict 60 control-decision values, but the closed-loop controller optimizes only 12 values (three blocks of four inputs). The implementation therefore retains the first 12 predicted values and computes a feasible initial decision:
\begin{equation}
z_k^{(0)}=\Pi_{\mathcal Z}\!\left(\pi_\theta(x_k)_{1:12}\right),\quad
\Pi_{\mathcal Z}(v)=\arg\min_{z\in\mathcal Z}\tfrac12\|z-v\|_2^2.
\end{equation}
The projection imposes the implemented input-decision bounds; it does not guarantee state-trajectory safety. The 60-to-12 reduction is an implementation limitation discussed in Section~5.

\section{Transformer training and use in MPC}
\subsection{Why the Transformer was introduced}
We first tested a Transformer as a predictor of future UAV motion. In that approach, the network would replace repeated physical simulations inside MPC. The early closed-loop tests were not sufficiently reliable, and restricting rotor commands to a narrow range around hover improved their behavior. We therefore changed the question: can a Transformer provide a useful starting point for the existing physics-based optimizer? In the experiments presented in this paper, the mechanical model is used to predict the motion of the UAV and its suspended payload during MPC optimization. The Transformer provides an initial guess for the control inputs, which is subsequently refined by the optimizer. It does not replace the mechanical model or calculate the predicted motion.

\subsection{Training data and label quality}
The Transformer was trained using 200 examples obtained from earlier SCP optimizations: eight initial examples, eight additional examples, and 184 further examples. Each example contains a simulated UAV state with 22 variables and a corresponding control sequence with 60 values calculated by the SCP solver.

Of the 184 further examples, 132 were recorded as converged. The remaining 52 were accepted as successful solver runs, although convergence was not confirmed. Thus, the training data consist of solutions produced by the numerical optimizer, but they are not necessarily globally optimal.

The data were checked for invalid numerical values. However, the separation between the training, validation, and test datasets was not fully verified for possible overlap. Consequently, we do not claim that these datasets are completely independent.

\subsection{Transformer input, architecture, and training}
The Transformer receives the current UAV state, represented by 22 variables. After normalization, this state is repeated six times, with an additional position value between $-1$ and $1$. The resulting input has dimensions $23\times6$. These six positions do not represent successive measurements in time; they are used only to organize the input to the network.

The network maps the input to 32 features and processes them using two-head self-attention with 16 key channels. The resulting features are normalized and averaged across the six positions. A fully connected layer with 64 neurons and a ReLU activation then produces the final output of 60 control-decision values.

The network was trained using the Adam optimizer, a learning rate of $5\times10^{-4}$, minibatches of four, and a maximum of 120 epochs. The network with the lowest validation loss was retained. Input normalization was calculated from the training data, with a minimum standard deviation of 0.05 for each component. Training was configured to run on a CPU.

\subsection{How the prediction is used in the controller}
During training, the Transformer learns to predict 60 control values, representing 15 successive control blocks with four adjustments in each block. However, the MPC controller used in the final experiments employs only three control blocks, corresponding to 12 optimization variables.

For each MPC update, the controller takes the first 12 values predicted by the Transformer as its initial guess. If necessary, these values are adjusted to satisfy the imposed input constraints before optimization begins. The SCP optimizer then improves this initial guess using the same objective function and constraints as the other initialization methods.

This difference between training and implementation is a limitation of the present study. The Transformer is trained to predict a longer control sequence than the one used by the final controller. Training the network directly for the controller's 12 optimization variables would provide a more consistent approach and could be investigated in future work. The present experiments do not establish how much this mismatch affected performance.

\subsection{Feasibility projection}
For a shifted or Transformer-predicted initial vector $v$, the controller solves
\begin{equation}
\min_z\ \tfrac12\|z-v\|_2^2\quad\text{subject to}\quad Az\leq b,\quad -1\leq z_i\leq1.
\end{equation}
This small quadratic program is solved with a convex quadratic-programming solver. It enforces the implemented control-decision and rotor limits before SCP begins; it does not establish state-trajectory safety. Zero initialization already satisfies these limits and does not need projection. Reported computation times include Transformer prediction and projection when applicable.

\section{Experimental design}\label{sec:experimental}
The benchmark with independently solved MPC problems tests 12 held-out UAV states, comparing zero initialization with two Transformer models. For the closed-loop comparison, all three initialization methods use the same initial conditions, reference, mechanical model, prediction horizon, input constraints, and solver settings. The only difference at the start of each MPC optimization is the initial guess. Each method then applies its own optimized control action to a separate simulation and uses the resulting state at the next update. The trajectories may therefore differ slightly after the common starting point. An episode is one continuous simulated run of the UAV--payload system, starting from a specified initial state and lasting eight consecutive MPC updates (1.6~s). We test 12 such episodes with different initial conditions, for a total of 96 updates per method. The measured total time is
\begin{equation}
T_{\rm total}=T_{\rm inference}+T_{\rm projection}+T_{\rm SCP}.
\end{equation}
Physical plant propagation is excluded from this timer. All methods use the same physical predictor, residual, constraints, trust-region and SCP iteration budget. These tests are computational comparisons rather than hardware experiments.

\section{Results}
\begin{table}[ht]\centering\scriptsize\caption{Benchmark with independently solved MPC problems: 12 held-out optimization problems. Time and SCP iterations are means per optimization.}\label{tab:independent}\begin{tabular}{@{}lrr@{}}\toprule
Initialization & Mean time (s) & Mean SCP iterations\\\midrule
Zero & 5.268 & 3.000\\
Transformer (16) & 4.776 & 2.670\\
Transformer (200) & 4.345 & 2.500\\\bottomrule\end{tabular}\end{table}

\begin{table*}[t]\centering\caption{Closed-loop benchmark: 12 held-out episodes and 96 MPC updates per method. Time and SCP iterations are means per MPC update. Valid counts updates satisfying the input constraints and improving the objective; Converged counts updates satisfying the SCP stopping test.}\label{tab:closed}\begin{tabular}{@{}lrrrr@{}}\toprule
Initialization & Mean time (s) & Mean SCP iterations & Valid & Converged\\\midrule
Zero & 5.059 & 3.000 & 96/96 & 96/96\\
Shift & \textbf{4.019} & \textbf{2.354} & 96/96 & 96/96\\
Transformer (200 labels) & 4.802 & 2.844 & 96/96 & 95/96\\\bottomrule\end{tabular}\end{table*}

Table~\ref{tab:independent} reports the benchmark with independently solved MPC problems, in which 12 separate MPC optimization problems are solved without a preceding control solution. Transformer initialization trained with 200 examples reduced mean computation time from 5.268~s with zero initialization to 4.345~s, a reduction of 17.5\%. The mean number of SCP iterations decreased from 3.000 to 2.500.

Table~\ref{tab:closed} reports the closed-loop comparison. Each initialization method was tested over 12 episodes with eight consecutive MPC updates per episode. Shifted initialization gave the lowest mean computation time, 4.019~s per update, compared with 5.059~s for zero initialization and 4.802~s for Transformer initialization. Relative to zero initialization, these represent reductions of 20.6\% and 5.1\%, respectively. Transformer initialization was 19.5\% slower than shifted initialization. Shifted initialization was faster in 11 of the 12 episodes. The methods produced similar tracking results under the tested conditions, as illustrated below.

\subsection{Physical closed-loop trajectories}
\begin{figure*}[t]\centering
\includegraphics[width=.93\textwidth]{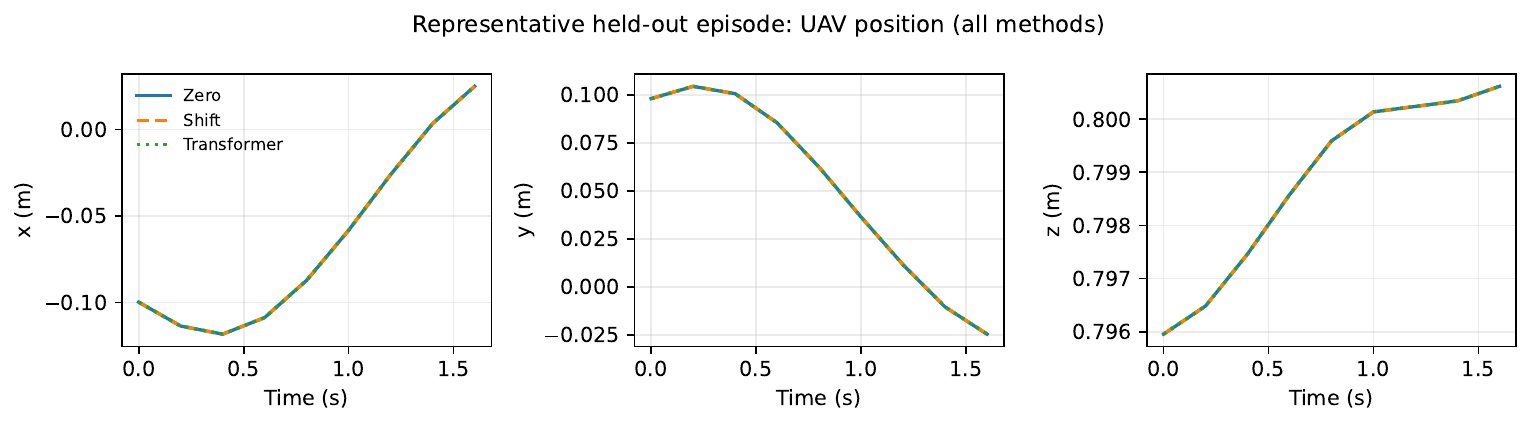}
\caption{Recorded quadrotor position components over the first held-out closed-loop episode. Zero, shifted, and Transformer initialization share the same initial condition, predictor, constraints, and SCP settings.}\label{fig:physical-position}\end{figure*}

\begin{figure}[t]\centering
\includegraphics[width=.98\columnwidth]{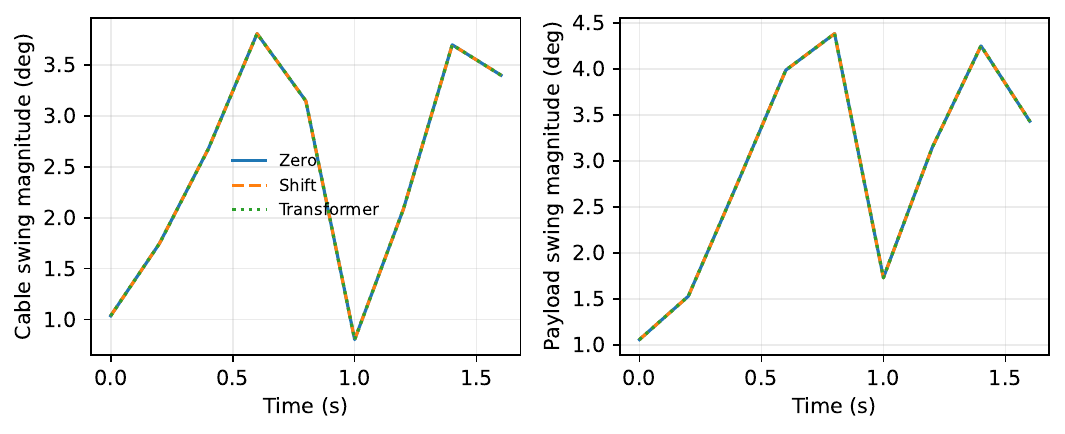}
\caption{Recorded magnitudes of the two swing-angle pairs over the same episode, calculated as $\sqrt{\alpha^2+\beta^2}$ and $\sqrt{\alpha_2^2+\beta_2^2}$ and converted to degrees. These are angle-vector norms, not exact spatial deflection angles.}\label{fig:physical-swing}\end{figure}

Figures~\ref{fig:physical-position} and~\ref{fig:physical-swing} show the simulated UAV position and cable and payload swing during the first test episode (1.6~s). All three methods start from the same initial state, but each follows the control actions calculated by its own MPC optimizer. This episode was selected by its order in the test data, not because it produced particularly favorable results.

\begin{figure}[t]\centering
\includegraphics[width=.98\columnwidth]{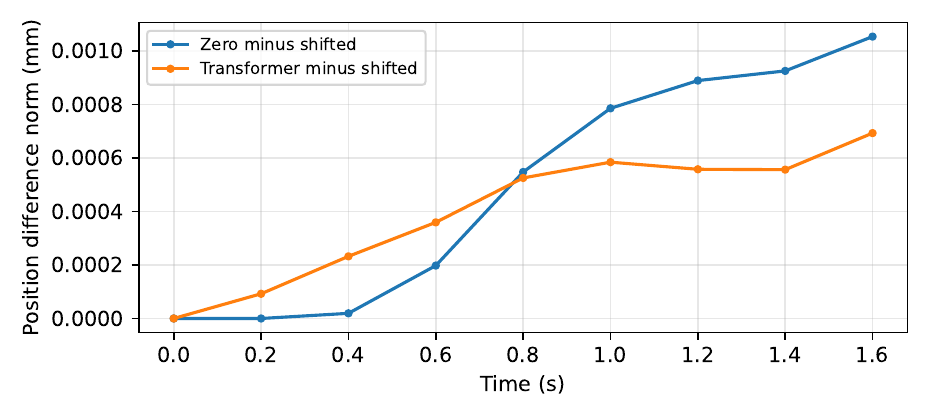}
\caption{First held-out episode: Euclidean position difference relative to the shifted-initialization trajectory (millimetres). The nonzero curves explain why the absolute position plots overlap; they are not a measure of landing or payload safety.}\label{fig:position-diff}\end{figure}

The position curves almost overlap at the scale shown in Fig.~\ref{fig:physical-position}. Figure~\ref{fig:position-diff} therefore shows their small differences relative to shifted initialization on a finer scale. The swing-angle plots provide a corresponding comparison for the suspended system. Similar motion during this episode does not establish identical control actions or comparable behavior under other conditions. These are numerical simulations, not flight measurements.

\begin{figure}[t]\centering
\includegraphics[width=.98\columnwidth]{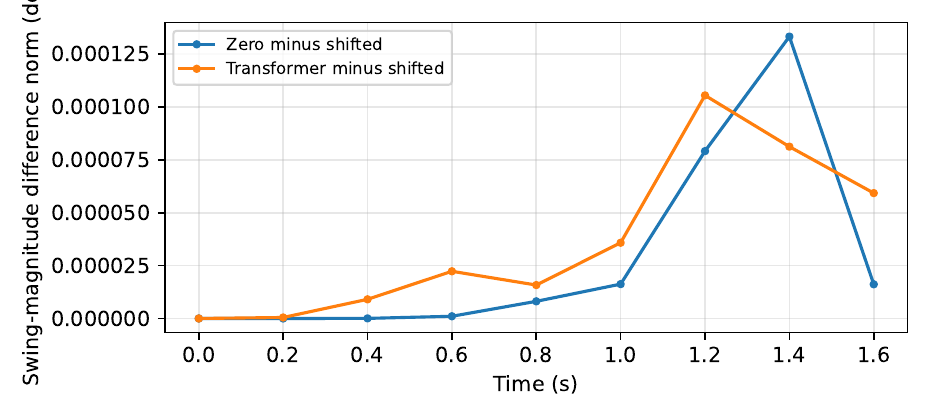}
\caption{First held-out episode: Euclidean difference of the two swing-angle magnitudes relative to the shifted-initialization trajectory (degrees). The scale is intentionally expanded and explicitly labelled.}\label{fig:swing-diff}\end{figure}
The differences in Figs.~\ref{fig:position-diff} and~\ref{fig:swing-diff} were calculated from the recorded 22-state trajectories for this episode, without running new simulations. They help reveal small differences between methods, but do not establish improved safety or tracking across all episodes.

\subsection{Computation time and SCP iterations across episodes}
\begin{figure}[ht]\centering\includegraphics[width=.97\linewidth]{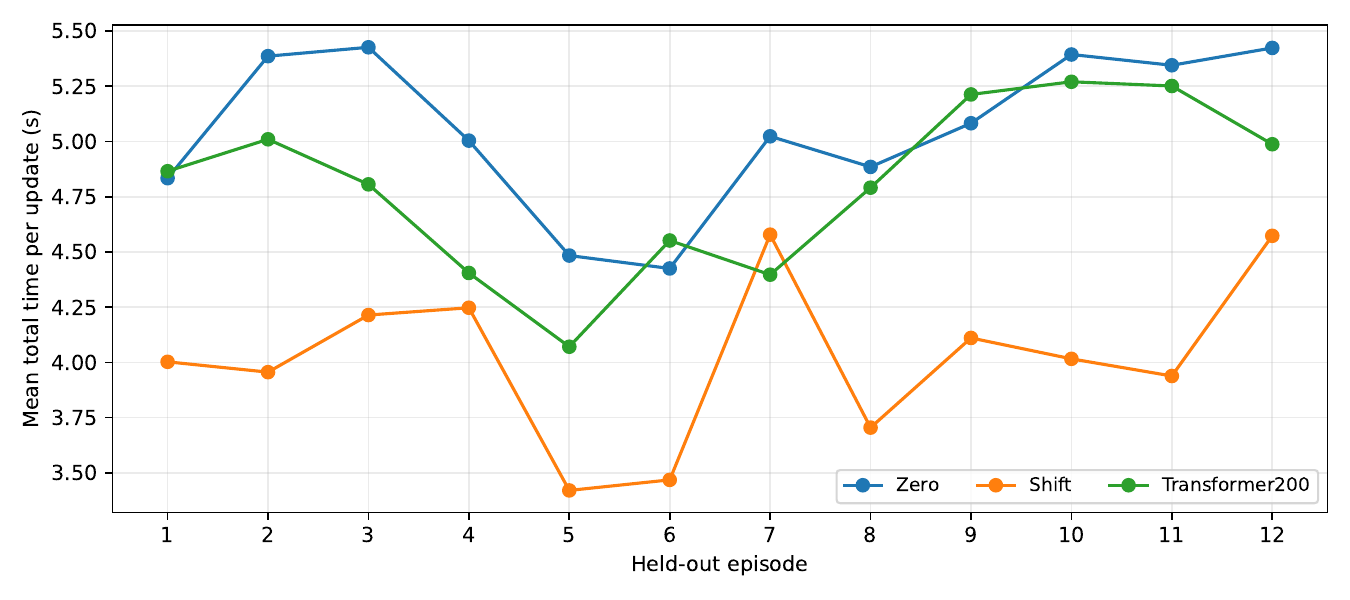}
\caption{Matched closed-loop benchmark episode-level mean total time per MPC update, measured as initialization inference plus projection plus SCP, excluding physical plant propagation.}\label{fig:episode-times}\end{figure}

\begin{figure}[ht]\centering\includegraphics[width=.97\linewidth]{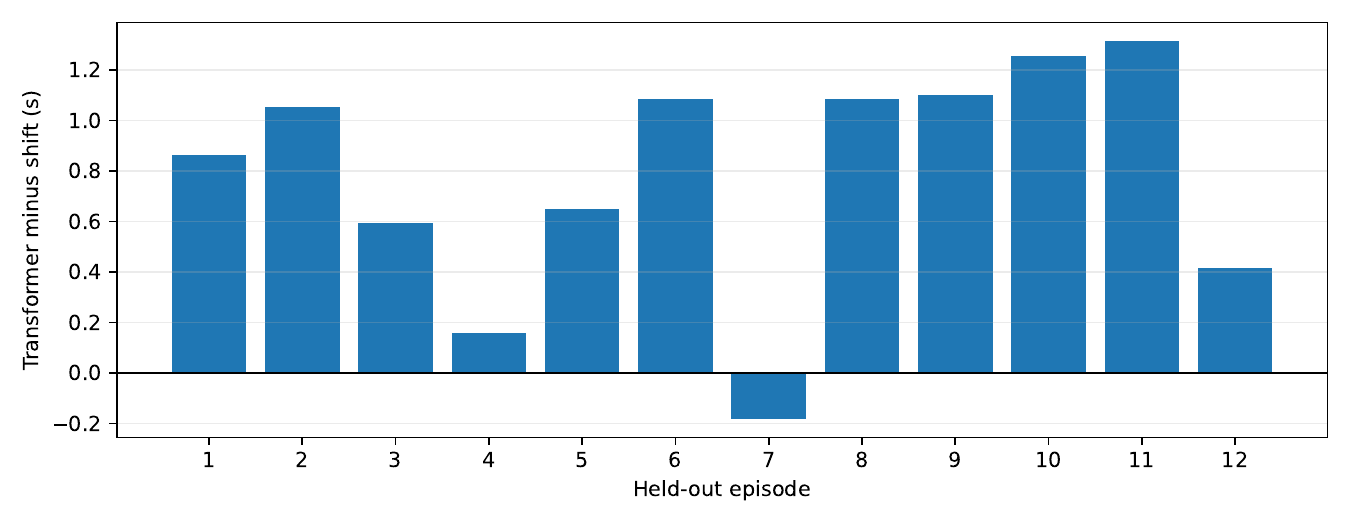}
\caption{Paired episode-level runtime difference: Transformer (200 labels) minus shift. Positive values mean that the Transformer approach was slower.}\label{fig:paired}\end{figure}

\begin{figure}[ht]\centering\includegraphics[width=.97\linewidth]{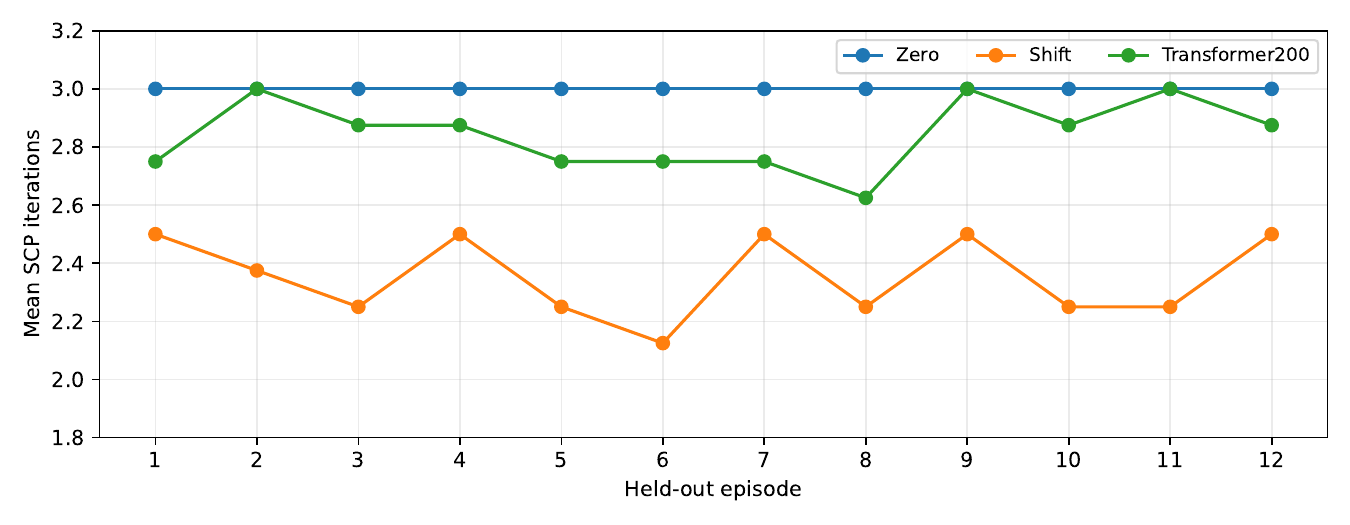}
\caption{Episode-level mean SCP iterations under the same four-iteration limit.}\label{fig:episode-iters}\end{figure}

Figures~\ref{fig:episode-times}--\ref{fig:episode-iters} show how computation time and SCP iteration counts vary across the 12 test episodes. Each plotted value is the average of eight consecutive MPC updates within one episode. The methods start each episode from the same state, but follow their own simulated trajectories after the first control action.
The difference between Transformer and shifted initialization was, on average, 0.782~s per update in favor of shifted initialization. The standard deviation of this difference across the 12 episodes was 0.465~s. These values describe variation in the present experiments, not a confidence interval for other flight conditions. The episode-level results show that the size of the computational advantage varies and that the faster method is not always the same.

\subsection{Separately solved problems and sequential closed-loop MPC}
The two benchmarks examine different situations. In the benchmark with independently solved MPC problems, each optimization problem is solved separately, so no solution from a previous MPC update is available for shifting. Here, Transformer initialization is faster than zero initialization. In the closed-loop benchmark, the controller solves successive, related optimization problems. Shifted initialization can reuse the preceding solution and is faster than Transformer initialization in the tested episodes. The findings are consistent because the benchmarks test different situations.

\subsection{Computation time and real-time requirements}
The reported time per MPC update, $T_{\rm total}$, includes generating the initial control guess, adjusting that guess to satisfy input constraints when necessary, and carrying out the SCP optimization, as defined in Section~\ref{sec:experimental}. The time required to advance the simulated UAV--payload system is not included. The physical system was propagated separately using an adaptive fourth/fifth-order Runge--Kutta integrator.

The MPC sampling interval is $T_s=0.2$~s, but the measured computation times are several seconds per update. Thus, none of the three methods meets the real-time requirement of the present controller configuration. These experiments compare computational performance; they do not demonstrate readiness for real-time flight.

\section{Methodological precursor: Python reproduction of Transformer-assisted SCP}
Before the UAV warm-start study, a separate Python notebook in Google Colab examined the Transformer-assisted trajectory-optimization approach described in the cited work. This was a preliminary reproduction study, not a claim of exact reproduction of the authors' original software or all their published experiments. The robot, obstacle geometry, and solver used in the notebook differ from those of the cable-suspended UAV studied here. In one recorded test, using a Transformer-generated initial guess reduced the SCP optimization time from 6.277~s to 3.561~s, with nearly equal final objective values (approximately 0.0291121). However, generating the initial guess with the trained Transformer (Transformer inference) required another 14.916~s. The combined time was therefore 18.477~s, longer than the conventional SCP solve. In this test, the Transformer reduced the optimization time but did not reduce the overall computation time.

A learned candidate also exhibited an obstacle-constraint violation before SCP refinement. This motivated explicit distinction between candidate prediction, feasible initialization, and final optimized trajectory. The notebook should be archived as a separate reproduction record; neither its numerical timing nor its obstacle result belongs in the matched closed-loop benchmark statistical sample. The reproduction is not a verification of every experiment in the cited publication.

\section{Discussion}
The two experiments examine different MPC settings. In the first, each optimization problem is solved separately, without a previous control solution available for reuse. In the second, MPC problems are solved consecutively as the UAV--payload system evolves. Shifted initialization can therefore reuse the optimized control sequence from the preceding update. By contrast, the Transformer used here generates its initial guess from the current state without receiving that preceding solution. This difference helps explain why the Transformer improves on zero initialization for separate problems but does not outperform shifting in the sequential closed-loop comparison. The results do not establish that shifting is always preferable or that a different learned initializer could not perform better.

In the closed-loop comparison, all 96 MPC updates produced control decisions that passed the implemented validity checks for each method. However, one Transformer-initialized update did not satisfy the solver's recorded convergence test. A valid decision therefore does not necessarily mean that the optimization has converged. The reported computation times include generation of the initial guess, adjustment to satisfy input constraints when needed, and SCP optimization. All three methods required substantially more than the available 0.2~s per update.

The model represents the coupled motion of the UAV, elastic cable, and suspended payload, but does not include detailed rotor aerodynamics or atmospheric turbulence. The study in \cite{davoudi2020} illustrates how aerodynamic and wind assumptions can affect simulated quadrotor trajectories. The present results therefore do not establish performance under realistic wind disturbances. The guidance study in \cite{tsay2014} provides background on conventional quadrotor guidance and the conversion of control commands into individual rotor inputs, but does not examine a suspended payload or MPC warm starts. Neither reference validates the present controller.

The flight-validated predictive controller in \cite{rohr2021} operated at 20~Hz on a tiltwing UAV using a specialized implementation. Its aircraft, objectives, hardware, and numerical methods differ from ours, so its execution time is not directly comparable. In our study, each MPC optimization takes several seconds, although a new control action is required every 0.2~s. The measured times include numerical model integration carried out \emph{inside} the SCP optimization, but exclude the separate integration used to advance the simulated UAV--payload system after a control action is selected. The reported results therefore do not demonstrate real-time operation.

\subsection{Future work}
A first direction is to test a Transformer that improves the shifted previous solution rather than replacing it with a prediction based only on the current state. One possible formulation is
\begin{equation}
z_k^{(0)}=\Pi_{\mathcal Z}\bigl(z_k^{\rm shift}+\Delta_\theta(x_k,z_k^{\rm shift})\bigr),
\end{equation}
where the learned correction is projected onto the admissible input set. Training directly for the controller's 12 decision variables, and assessing whether such a predictor reduces total computation time, would address limitations identified here. Optimizer-aware fine-tuning is another possibility. None of these extensions has been tested in this study.

A second direction is to introduce wind disturbances and uncertain aerodynamic forces, together with measurements affected by noise and delay. A Kalman-filter-based estimator could be examined when a suitable stochastic model is available; interval observers or other set-membership estimators could instead be used to represent bounded uncertainty. These approaches would require appropriate models and measurements and should be compared under the same disturbances and sensing conditions. Their effectiveness is not established by the present simulations.

Finally, experiments with instrumented, cable-suspended UAVs could help assess the model and controller under physical operating conditions. Possible measurements include UAV position and attitude, cable tension or extension, and payload motion. Fiber-optic sensing along an instrumented cable could be explored for strain or deformation measurements, subject to sensor packaging, calibration, weight, and data-acquisition constraints. Such sensing should not be confused with the communication tether used by some tethered drones: a fiber-optic communication link alone does not measure cable deformation. Whether distributed fiber-optic sensing is practical for the short, moving suspension studied here remains an open experimental question. Before field deployment, the controller would also need substantially faster computation and tests of estimation, robustness, and safety. The earlier Colab reproduction remains a separate methodological record, not additional UAV flight evidence.

\section{Conclusions}
The study provides a controlled numerical comparison of zero, shifted, and Transformer-based initializations within a common sequential convex programming (SCP) model predictive controller for a multirotor UAV with a cable-suspended payload. The principal findings are:
\begin{enumerate}
\item \textbf{Predictive-control framework.} The implemented MPC formulation predicts the coupled UAV--payload motion over a finite horizon, coordinates the control inputs, and explicitly imposes the modeled input constraints. Holding this controller fixed across the experiments makes it possible to isolate the effect of initialization. These are useful features of the formulation, but the experiments do not establish superiority over a non-MPC controller or guarantee physical safety.
\item \textbf{Benefit of learned initialization.} The Transformer trained on 200 expert-labeled records improves initialization relative to zero in the tested comparisons. In the matched 12-episode closed-loop benchmark, its mean update time is 4.802~s, compared with 5.059~s for zero initialization, a reduction of approximately 5.1\%.
\item \textbf{Advantage of the shifted baseline.} Conventional shifted initialization gives the lowest mean update time (4.019~s) and the fewest mean SCP iterations (2.354, versus 2.844 for the Transformer and 3.000 for zero initialization). The learned initializer therefore does not outperform this inexpensive receding-horizon baseline in the reported benchmark.
\item \textbf{Computational and safety limitations.} None of the tested strategies meets the 0.20-s control-update deadline. Input feasibility and numerical validity do not imply collision avoidance, cable-tension safety, robustness, closed-loop stability, or recursive feasibility. The findings are limited to the investigated model, training data, horizon, constraints, and solver configuration.
\end{enumerate}

A positive outcome is the establishment of a consistent numerical framework in which a mechanically coupled UAV--payload model, constrained predictive control, and alternative initialization strategies can be evaluated without changing the underlying controller. The MPC formulation allows anticipated payload motion and actuator limits to enter the optimization explicitly, while the matched experiments show that initialization choice has a measurable effect on computational effort. The Transformer results demonstrate that learned initialization can improve on a zero baseline, even though it does not surpass shifting in this case. This combination of a clearly specified comparison protocol and a clear negative result against a strong conventional baseline is useful for deciding when learned warm starts warrant their additional training and inference costs. The present simulations do not establish that MPC is superior to other control architectures; that question would require a separate matched controller comparison.

\section*{Code and data availability}
The source code and numerical datasets are not publicly available with this preprint. The numerical results are reported from archived simulation records; independent recomputation of the published summary statistics requires access to those records and the corresponding numerical implementation. The accompanying manuscript source package contains the LaTeX and figure assets, but not the simulation code or underlying benchmark datasets. All reported results are numerical simulations; no hardware flight validation is claimed.

\appendix
\section{Shifted initialization with fixed move blocking}
Let $Z=[a,b,c]$, where each block belongs to $\mathbb R^4$ and occupies two sampling intervals. With the expansion operator $E(Z)=[a,a,b,b,c,c]$, the one-sample shift and terminal hold produce $S(E(Z))=[a,b,b,c,c,c]$. The fixed-grid reblocking operator averages successive pairs, giving
\begin{equation}
R(S(E(Z)))=\left[\tfrac12(a+b),\tfrac12(b+c),c\right].
\end{equation}
This candidate is projected onto the same polyhedral admissible set used by the learned initialization. This appendix states the exact algorithmic convention; it is not a claim that shifting changes the move-block boundaries.

\end{document}